\RequirePackage[2020-02-02]{latexrelease}
\documentclass{rQUF2e}

\usepackage{epstopdf}% To incorporate .eps illustrations using PDFLaTeX, etc.
\usepackage{subfigure}% Support for small, `sub' figures and tables

\theoremstyle{plain}

\theoremstyle{definition}

\theoremstyle{remark}

\begin{document}

%\jvol{00} \jnum{00} \jyear{2014} \jmonth{October}

\title{Neural Generalised AutoRegressive Conditional Heteroskedasticity}

\author{Zexuan Yin$^{\ast}$$\dag$\thanks{$^\ast$Corresponding author.
Email: zexuan.yin.20@ucl.ac.uk.co.uk} and Paolo Barucca${\dag \ddag}$\\
\affil{$\dag$Deparment of Computer Science, University College London, WC1E 7JE, United Kingdom\\
$\ddagger$p.barucca@ucl.ac.uk} \received{v1.1 released November 2021} }

\maketitle

\begin{abstract}
We propose Neural GARCH, a class of methods to model conditional heteroskedasticity in financial time series. Neural GARCH is a neural network adaptation of the GARCH(1,1) model in the univariate case, and the diagonal BEKK(1,1) model in the multivariate case. We allow the coefficients of a GARCH model to be time-varying in order to reflect the constantly changing dynamics of financial markets. The time-varying coefficients are parameterised by a recurrent neural network that is trained with stochastic gradient variational Bayes. We propose two variants of our model, one with normal innovations and the other with Student's t innovations. We test our models on a wide range of univariate and multivariate financial time series, and we find that the Neural Student's t model consistently outperforms the others.
\end{abstract}


\begin{classcode}C32, C45, C53, \end{classcode}

\section{Introduction}

Modelling conditional heteroskedasticity (time-varying volatility) in financial time series such as energy prices \citep{Chan2016}, cryptocurrencies \citep{Chu2017}, and foreign currency exchange rates \cite{Malik2005} is of great importance to financial practitioners as it allows better decision making with regards to portfolio selection, asset pricing and risk management. In the univariate setting, popular methods include Autoregressive Conditional Heteroskedastic models (ARCH) \citep{Engle1982} and Generalised GARCH (GARCH) models \citep{Bollerslev1986}. ARCH and GARCH models are regression-based models estimated using maximum likelihood, and are capable of capturing stylised facts about financial time series such as volatility clustering \citep{Bauwens2006}. The ARCH($p$) model describes the conditional volatility as a function of $p$ lagged squared residuals, and similarly the GARCH($p$,$q$) model include contributions due to the last $q$ conditional variances. Many variants of the GARCH model have been proposed to better capture properties of financial time series, for example the EGARCH \citep{Nelson1991} and GJR-GARCH \citep{GLOSTEN1993} models were designed to capture the so-called leverage effect, which describes the negative relationship between asset price and volatility.

In a multivariate setting, instead of modelling only time-varying conditional variances, for an $n$-dimensional system, we estimate the $n\times n$ time-varying variance-covariance matrix. This allows us to investigate interactions between the volatility of different time series and whether there is a transmission of volatility (spillover effect) between markets \citep{Bauwens2006,Erten2012}. Popular multivariate GARCH models include the VEC model \citep{Bollerslev1988}, the BEKK model \citep{Engle1995}, the GO-GARCH model \citep{VanDerWeide2002} and DCC model \citep{Christodoulakis2002,Tse2002,Engle2002}. 

In this paper we focus specifically on GARCH(1,1) models in the univariate case and the diagonal BEKK(1,1) model in the multivariate case to model daily financial asset returns. We consider several assets classes such as foreign exchange rates, commodities and stock indices. GARCH(1,1) models work well in general practical settings due to their simplicity and robustness to overfitting \citep{Wu2013}. 

In traditional GARCH models, the estimated coefficients are constant which imply a stationary returns process with a constant unconditional mean and variance \citep{Bollerslev1986}. However, there is evidence in existing literature that relaxing the stationary constraint on the returns time series can often lead to a better performance as it allows the model to better capture time-varying market conditions. In \cite{Starica2005} the authors modelled daily S\&P 500 returns with locally stationary models and found that most of the dynamics were concentrated in shifts of the unconditional variance, and forecasts based on non-stationary unconditional modelling yielded a better performance than a stationary GARCH(1,1) model. Similarly, the authors in \cite{Wu2013} designed a GARCH(1,1) model with time-varying coefficients that followed a random walk process, and they reported better forecasting performances in the test dataset relative to the GARCH(1,1) model. 

To this end, we propose univariate and multivariate GARCH models with time-varying coefficients that are parameterised by a recurrent neural network. Our method allows the simplicity and interpretability of GARCH models to be combined with the expressive power of neural networks, and this approach follows a trend in the literature that combines classical time series models with deep learning. In \cite{Rangapuram2018} for example, the authors proposed to parameterise the coefficients of a linear Gaussian state space model with a recurrent neural network, and the latent states were then inferred using a Kalman filter. This approach is advantageous as the neural network allows modelling of more complex relationships between time steps whilst preserving the structural form of the state space model. Similarly, by preserving the structural form of the BEKK model, we can obtain covariance matrices that are symmetric and positive definite \citep{Engle1995} without the need of implementing further constraints. We treat the time-varying GARCH coefficients as latent variables to be inferred, and to achieve this we leverage recent advances in amortised variational inference in the form of a variational autoencoder (VAE) \citep{Kingma2014}, and subsequent combinations of a VAE with a recurrent neural network (so-called Variational RNN, or VRNN) \citep{Chung2015, Bayer2014,Krishnan2017,Fabius2015,Fraccaro2016,Karl2017} to allow efficient structured inference over a sequence of latent random variables.     

The rest of the paper is organised as follows: in Section \ref{sec: Preliminaries} we outline the preliminary mathematical concepts of GARCH modelling and amortised variational inference, in Section \ref{sec: Materials and Methods} we introduce the generative and inference model components of Neural GARCH, and in Section \ref{sec: Results} we present the performance of Neural GARCH on univariate and multivariate daily returns time series covering foreign exchange rates, commodity prices, and stock indices.

\section{Preliminaries}
\label{sec: Preliminaries}
\subsection{Univariate GARCH Model}
\label{sec: univariate garch}
The GARCH($p$,$q$) model \citep{Bollerslev1986} for a returns process $r_{t}$ is specified in terms of the conditional mean equation:
\begin{equation}
	r_{t} \sim \mathcal{N}(0,\sigma_{t}^2),
\end{equation}
and the conditional variance equation:
\begin{equation}
	\sigma_{t}^2 = \omega + \sum_{i=1}^{p} \alpha_{i}r_{t-i}^2 + \sum_{j=1}^{q} \beta_{j}\sigma_{t-j}^2.
\end{equation}
Under the GARCH(1,1) model, the returns process $r_{t}$ is covariance stationary with a constant unconditional mean and variance given by $\mathbb{E}[r_{t}]=0$ and $\mathbb{E}[r_{t}^2]=\frac{\omega}{1-\alpha-\beta}$, where $\omega>0$, $\alpha\ge0$ and $\beta\ge0$ to ensure that $\sigma_{t}^2>0$, and $\alpha+\beta<1$ to ensure a finite unconditional variance.
For parameter estimation assuming normal innovations, the following log-likelihood function is maximised:
\begin{equation}
	\label{eqn: log normal}
	\mathcal{L} = -\sum_{t=1}^{T} (\frac{1}{2}log(\sigma_{t}^2) + \frac{r_{t}^2}{2\sigma_{t}^2})
\end{equation}

To model the leptokurtic (fat-tailed) behaviour of financial returns, the authors in \cite{Bollerslev1987} considered GARCH models with Student's t innovations with the following log-likelihood function to be maximised:
\begin{equation}
	\label{eqn: log student}
	\mathcal{L} = -\sum_{t=1}^{T} (log\Gamma(\frac{\nu+1}{2}) + log\Gamma(\frac{\nu}{2}) + \frac{1}{2}log(\nu-2) + \frac{1}{2}log(\sigma_{t}^2) + \frac{(\nu+1)}{2}log(1 + \frac{r_{t}^2}{(\nu-2)\sigma_{t}^2})),
\end{equation}
where $\nu>2$ is the degree of freedom and $\Gamma$ is the gamma function.

\subsection{BEKK Model}
The BEKK multivariate GARCH model \citep{Engle1995} parameterises an $n$-dimensional multivariate returns process $\boldsymbol{r}_{t}\in\mathbb{R}^{n\times T}$:
\begin{equation}
	\boldsymbol{r}_{t} \sim \mathcal{N}(0,\boldsymbol{\Sigma}_{t}),
\end{equation}
\begin{equation}
	\boldsymbol{\Sigma}_{t} = \boldsymbol{\Omega}^{T}\boldsymbol{\Omega} + \sum_{i=1}^{p} \boldsymbol{A}_{i}^{T}\boldsymbol{r}_{t-i}\boldsymbol{r}_{t-i}^{T}\boldsymbol{A}_{i} + \sum_{j=1}^{q} \boldsymbol{B}_{j}^{T}\boldsymbol{\Sigma}_{t-j}\boldsymbol{B}_{j},
\end{equation}
where $\boldsymbol{\Sigma}_{t}$ is the $n\times n$ symmetric and positive-definite covariance matrix, 
$\boldsymbol{\Omega}$ is an upper triangular matrix with $\frac{n(n+1)}{2}$ non-zero entries, $\boldsymbol{A}$ and $\boldsymbol{B}$ are $n\times n$ coefficient matrices. In our paper we consider the diagonal-BEKK model where $\boldsymbol{A}$ and $\boldsymbol{B}$ are diagonal matrices. 

\subsection{Neural Network Variational Inference}
For a latent variable model with parameters $\theta$, target variable $y$ and latent variable $z$, we wish to maximise the marginal likelihood with the latent variable integrated out, which often involves an intractable integral:
\begin{equation}
	logP_{\theta}(y) = log \int P_{\theta}(y|z)P_{\theta}(z)\,dz,
\end{equation}
instead we perform variable inference by approximating the actual posterior distribution $P_{\theta}(z|y)$ with a variational approxiation $q_{\phi}(z|y)$ and maximise the evidence lower bound ($ELBO$) where $logP_{\theta}(y) \ge ELBO$, which is equivalent to minimising the Kullback-Leiber ($KL$) divergence between the variational posterior $q_{\phi}(z|y)$ and the actual posterior $P_{\theta}(z|y)$ \citep{Kingma2014}:
\begin{equation}
	logP_{\theta}(y) = ELBO + KL(q_{\phi}(z|y)||P_{\theta}(z|y)),    
\end{equation}
where the $ELBO$ is given by:
\begin{equation}
	ELBO = \mathbb{E}_{z \sim q(z|x)} [logP_{\theta}(x|z)] - KL(q_{\phi}(z|x)||p_{\theta}(z)),
\end{equation}
where $P_{\theta}(z)$ is a prior distribution for $z$, and in a variational autoencoder (VAE) the generative and inference distributions $logP_{\theta}(x|z)$ and $q_{\phi}(z|x)$ are parameterised by neural networks. An uninformative prior such as $\mathcal{N}(0,1)$ is often used for the prior $P_{\theta}(z)$, however in our model we adopt a learned prior distribution $P_{\theta}(z|\mathcal{I}_{t-1})$ where $\mathcal{I}_{t-1}$ is the information set up to the time step $t-1$. This learned prior approach has achieved great success in sequential generation tasks such as video prediction \citep{Franceschi2020,Denton2018}.

\section{Materials and Methods}
\label{sec: Materials and Methods}
\subsection{Neural GARCH Models}
In this section we introduce the intuition and various components of Neural GARCH models. We shall focus specifically on univariate and multivariate GARCH(1,1) models as we would like to keep the GARCH model structure as simple as possible and delegate the modelling of complex relationships between time steps to the underlying neural network which outputs the coefficients of the GARCH models. For the rest of this paper, we use the terms (multivariate)GARCH(1,1) and BEKK(1,1) interchangeably when referring to multivariate systems.

In neural GARCH, the coefficients \{$\omega$, $\alpha$, $\beta$\} in the univariate case and \{$\boldsymbol{\Omega}$, $\boldsymbol{A}$, $\boldsymbol{B}$\} in the multivariate case are allowed to vary freely with time. This approach allows the model to capture the time-varying nature of market dynamics \cite{Wu2013}. The GARCH(1,1) and BEKK(1,1) models thus become:
\begin{equation}
	\label{eqn: neural garch}
	\sigma_{t}^2 = \omega_t + \alpha_tr_{t-1}^2 + \beta_t\sigma_{t-1}^2,   
\end{equation}
\begin{equation}
	\label{eqn: neural bekk}
	\boldsymbol{\Sigma}_{t} = \boldsymbol{\Omega}^{T}_t\boldsymbol{\Omega}_t + \boldsymbol{A}_{t}^{T}\boldsymbol{r}_{t-1}\boldsymbol{r}_{t-1}^{T}\boldsymbol{A}_{t} +  \boldsymbol{B}_{t}^{T}\boldsymbol{\Sigma}_{t-1}\boldsymbol{B}_{t},
\end{equation}
For notation purposes we define the parameter set $\boldsymbol{\gamma}_t=[\omega_t, \alpha_t, \beta_t]^T$ for GARCH(1,1) and $\boldsymbol{\gamma}_t=[\boldsymbol{\Omega}_t, \boldsymbol{A}_t, \boldsymbol{B}_t]^T$ for BEKK(1,1). 

In our proposed framework, $\boldsymbol{\gamma}_t$ is a multivariate normal latent random variable with a diagonal covariance matrix to be estimated at every time step. For GARCH(1,1) this involves an estimation of a vector of size 3 for a model with normal innovations: 
\begin{equation}
	\boldsymbol{\gamma}_t = \begin{bmatrix} \omega_t \\ \alpha_t \\ \beta_t \end{bmatrix} \sim \mathcal{N}(\boldsymbol{\mu_t},\boldsymbol{\Sigma}_{\gamma,t}),
\end{equation}
and the vector $[\sigma_{\omega_t}^2,\sigma_{\alpha_t}^2,\sigma_{\beta_t}^2]^T$ represents the diagonal elements of the covariance matrix $\boldsymbol{\Sigma}_{\gamma,t}$. Here we have written the covariance matrix of the parameter set $\boldsymbol{\gamma}_t$ as $\boldsymbol{\Sigma}_{\gamma,t}$ in order to distinguish it from the covariance matrix of the asset returns $\boldsymbol{\Sigma}_t$. For neural GARCH(1,1) with Student's t innovations, $\boldsymbol{\gamma}_t$ is augmented with the degree of freedom parameter $\nu_t$ such that $\boldsymbol{\gamma}_t = [\omega_t, \alpha_t, \beta_t, \nu_t]^T$.

For the multivariate diagonal BEKK(1,1), we adopt a similarly methodology. For a system of $n$ assets, $\boldsymbol{\gamma}_t$ of a model with normal innovations is a vector of size $2n + \frac{n(n+1)}{2}$ \citep{Engle1995}, and with Student's t innovations $\boldsymbol{\gamma}_t$ is of size $2n + \frac{n(n+1)}{2} + 1$. As an example, for a system of 2 assets ($n=2$), the BEKK model is given by:

\begin{multline}
	\boldsymbol{\Sigma}_t = \begin{bmatrix} c_{11,t} & 0 \\ c_{21,t} & c_{22,t} \end{bmatrix}\begin{bmatrix} c_{11,t} & c_{12,t} \\ 0 & c_{22,t} \end{bmatrix} + \begin{bmatrix} a_{11,t} & 0 \\ 0 & a_{22,t} \end{bmatrix}\begin{bmatrix} r_{1,t-1} \\ r_{2,t-1} \end{bmatrix}\begin{bmatrix} r_{1,t-1} \\ r_{2,t-1} \end{bmatrix}^T\begin{bmatrix} a_{11,t} & 0 \\ 0 & a_{22,t} \end{bmatrix}\\ + \begin{bmatrix} b_{11,t} & 0 \\ 0 & b_{22,t} \end{bmatrix}\begin{bmatrix} \sigma_{11,t}^2 & \sigma_{12,t}^2 \\ \sigma_{21,t}^2 & \sigma_{22,t}^2 \end{bmatrix}\begin{bmatrix} b_{11,t} & 0 \\ 0 & b_{22,t} \end{bmatrix},
\end{multline}
where $a_{ij,t}$ is the $i,j$th element of the matrix $\boldsymbol{A}_t$, the parameter set $\boldsymbol{\gamma}_t$, which also has a multivariate normal distribution, is given by:
\begin{equation}
	\label{eqn: gamma}
	\boldsymbol{\gamma}_t = [a_{11,t}, a_{22,t}, b_{11,t}, b_{22,t}, c_{11,t}, c_{12,t}, c_{22,t}]^T
\end{equation}

The main contribution of our paper is the estimation of $\boldsymbol{\gamma}_t$ with a recurrent neural network (RNN) and a multilayer perceptron (MLP). We provide the exact estimation schemes in Sections \ref{sec: generative} and \ref{sec: inference}. Since we assume a multivariate normal distribution with a diagonal covariance matrix for $\boldsymbol{\gamma}_t$, we need to estimate the means and variances of the elements in $\boldsymbol{\gamma}_t$ with our neural network.

\subsection{Generative Model}
\label{sec: generative}
\begin{figure}[h]
	\centering
	\includegraphics[width=0.75\columnwidth,clip,keepaspectratio]{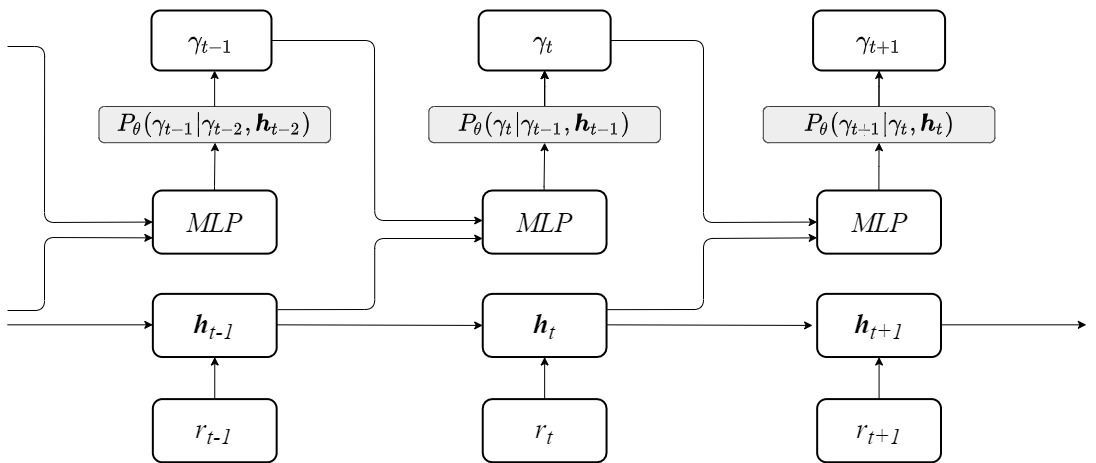}
	\caption{Generative model of neural GARCH. The generative MLP takes as input $\{\boldsymbol{\gamma}_{t-1}, \boldsymbol{h}_{t-1}\}$ and outputs the estimated means and variances of the elements in $\boldsymbol{\gamma}_t$.}
	\label{fig1}
\end{figure}

The generative model distribution $P_\theta(\boldsymbol{r}_{1:T}, \boldsymbol{\Sigma}_{1:T},\boldsymbol{\gamma}_{1:T})$ of a general multivariate neural GARCH is presented in Figure \ref{fig1} and given by (\ref{eqn: joint}). For the univariate case, one simply replaces $\boldsymbol{\Sigma}_t$ in (\ref{eqn: joint}) with $\sigma^2_t$.
\begin{equation}
	\label{eqn: joint}
	P_\theta(\boldsymbol{r}_{1:T}, \boldsymbol{\Sigma}_{1:T},\boldsymbol{\gamma}_{1:T}) = P(\boldsymbol{\gamma}_0)P(\boldsymbol{\Sigma}_0)\prod_{t=1}^{T}P_\theta(\boldsymbol{r}_t|\boldsymbol{\Sigma}_t)P_\theta(\boldsymbol{\Sigma}_t|\boldsymbol{\gamma}_t, \boldsymbol{r}_{t-1},\boldsymbol{\Sigma}_{t-1})P_\theta(\boldsymbol{\gamma}_t|\boldsymbol{\gamma}_{t-1},\boldsymbol{r}_{1:t-1}).
\end{equation}
The initial priors were set to delta distributions, $P(\boldsymbol{\Sigma}_0)$ was centered on the covariance matrix estimated using the training dataset, and $P(\boldsymbol{\gamma}_0)$ was centered on a vector of 1s. The predictive distribution $P_\theta(\boldsymbol{\gamma}_t|\boldsymbol{\gamma}_{t-1},\boldsymbol{r}_{1:t-1})$ takes as input the information set $\mathcal{I}_{t-1} = \{\boldsymbol{\gamma}_{t-1}, \boldsymbol{r}_{1:t-1}\}$ and predicts the 1-step-ahead value $\boldsymbol{\gamma}_t$. For this parameterisation, we leverage a recurrent neural network to carry $\boldsymbol{r}_{1:t-1}$ such that:
\begin{equation}
	P_\theta(\boldsymbol{\gamma}_t|\boldsymbol{\gamma}_{t-1},\boldsymbol{r}_{1:t-1}) = P_\theta(\boldsymbol{\gamma}_t|\boldsymbol{\gamma}_{t-1},\boldsymbol{h}_{t-1}),    
\end{equation}
where $\boldsymbol{h}_t$ is the hidden state of the underlying RNN. In our model we use a gated recurrent unit (GRU) \cite{Cho2014}. We then use an MLP which takes as input $\mathcal{I}_{t-1}$ as maps it to the means and variances of the elements in $\boldsymbol{\gamma}_t$. In the 2-dimensional example given in (\ref{eqn: gamma}), the estimation is done using:
\begin{equation}
	\label{eqn: prediction}
	[\mu_{a_{11,t}}, ..., \mu_{c_{22,t}}, \sigma^2_{a_{11,t}}, ..., \sigma^2_{c_{22,t}}]^T = MLP_{pred}(\boldsymbol{\gamma}_{t-1}, \boldsymbol{h}_{t-1}),
\end{equation}
and we apply a sigmoid function on the neural network output to ensure that the estimated variances of the elements in $\boldsymbol{\gamma}_t$ and the GARCH coefficients are non-negative. We have also tested other ways to ensure non-negativity such as using a softplus function, however we found that applying a sigmoid function gave the best performance. For neural GARCH with Student's t innovations, we require that $\nu>2$ in order to have a well-defined covariance. Since appyling the sigmoid function ensures our estimated coefficients are non-negative, we estimate $\nu' = \nu-2$ (instead of $\nu$ directly) to ensure $\nu>2$.

The conditional distribution $P_\theta(\boldsymbol{\Sigma}_t|\boldsymbol{\gamma}_t, \boldsymbol{r}_{t-1},\boldsymbol{\Sigma}_{t-1})$ is a delta distribution centered on (\ref{eqn: neural garch}) in the univariate case and (\ref{eqn: neural bekk}) in the multivariate case as we can calculate the covariance matrix $\boldsymbol{\Sigma_t}$ deterministically given $\{\boldsymbol{\gamma}_t, \boldsymbol{r}_{t-1}, \boldsymbol{\Sigma}_{t-1}\}$. The distribution $P_\theta(\boldsymbol{r}_t|\boldsymbol{\Sigma}_t)$ is the likelihood function and we have provided their logarithms (in the univariate case) in (\ref{eqn: log normal}) for normal innovations and (\ref{eqn: log student}) for Student's t innovations.

\subsection{Inference Model}
\label{sec: inference}
\begin{figure}[h]
	\centering
	\includegraphics[width=0.75\columnwidth,clip,keepaspectratio]{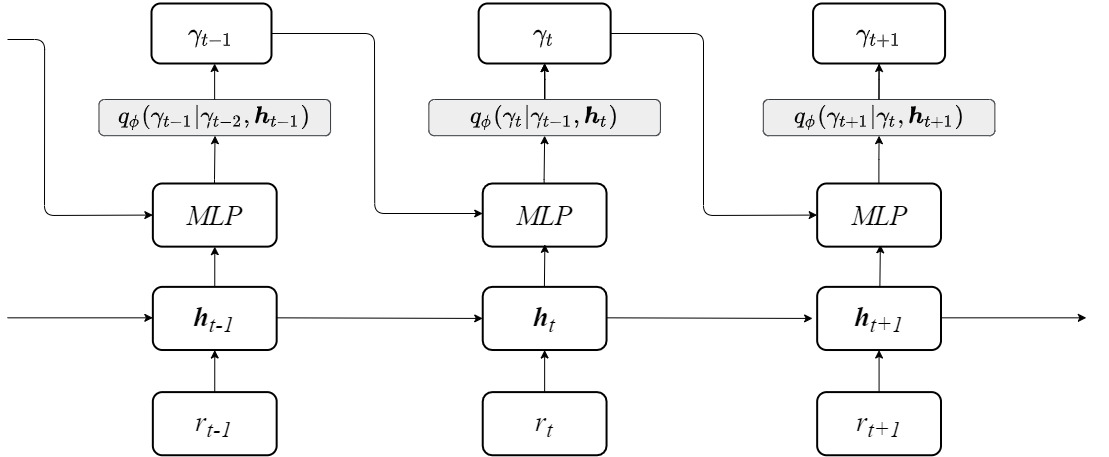}
	\caption{Inference model of neural GARCH. The inference MLP outputs the posterior estimate of $\boldsymbol{\gamma}_t$ conditioned on available information up to time $t$.}
	\label{fig2}
\end{figure}

The inference model distribution $q_\phi(\boldsymbol{\Sigma}_{1:T},\boldsymbol{\gamma}_{1:T}|\boldsymbol{r}_{1:T})$ is presented in Figure \ref{fig2} and can be factorised as: 
\begin{equation}
	\label{eqn: joint inf}
	q_\phi(\boldsymbol{\Sigma}_{1:T},\boldsymbol{\gamma}_{1:T}|\boldsymbol{r}_{1:T}) = P(\boldsymbol{\gamma}_0)P(\boldsymbol{\Sigma}_0)\prod_{t=1}^{T}q_\phi(\boldsymbol{\Sigma}_t|\boldsymbol{\gamma}_t, \boldsymbol{r}_{t-1},\boldsymbol{\Sigma}_{t-1})q_\phi(\boldsymbol{\gamma}_t|\boldsymbol{\gamma}_{t-1},\boldsymbol{r}_{1:t}),
\end{equation}
where $P(\boldsymbol{\gamma}_0)$ and $P(\boldsymbol{\Sigma}_0)$ are the same as in the generative model, $q_\phi(\boldsymbol{\Sigma}_t|\boldsymbol{\gamma}_t, \boldsymbol{r}_{t-1},\boldsymbol{\Sigma}_{t-1})$ has the same functional form (a delta distribution) as $P_\theta(\boldsymbol{\Sigma}_t|\boldsymbol{\gamma}_t, \boldsymbol{r}_{t-1},\boldsymbol{\Sigma}_{t-1})$, however $\boldsymbol{\gamma}_t$ is now drawn from the posterior distribution $q_\phi(\boldsymbol{\gamma}_t|\boldsymbol{\gamma}_{t-1},\boldsymbol{r}_{1:t})$ where:
\begin{equation}
	q_\phi(\boldsymbol{\gamma}_t|\boldsymbol{\gamma}_{t-1},\boldsymbol{r}_{1:t}) = q_\phi(\boldsymbol{\gamma}_t|\boldsymbol{\gamma}_{t-1},\boldsymbol{h}_{t}).
\end{equation}
We note that the generative and inference networks share the same underlying recurrent neural network but uses information at different time steps. The generative model predicts $\boldsymbol{\gamma}_t$ using the information set $\mathcal{I}_{t-1}$ and the inference model infers $\boldsymbol{\gamma}_t$ using $\mathcal{I}_t$. The inference MLP ($MLP_{inf}$) however is different to that of the generative model ($MLP_{pred}$) and it outputs the posterior estimates of the elements of $\boldsymbol{\gamma}_t$:
\begin{equation}
	\label{eqn: inference}
	[\mu_{a_{11,t}}, ..., \mu_{c_{22,t}}, \sigma^2_{a_{11,t}}, ..., \sigma^2_{c_{22,t}}]_{post}^T = MLP_{inf}(\boldsymbol{\gamma}_{t-1}, \boldsymbol{h}_{t}).
\end{equation}
\subsection{Model Training}
For neural network training we optimise the generative and inference model parameters ($\theta$ and $\phi$) jointly using stochastic gradient variational Bayes \cite{Kingma2014}. Our objective function is the ELBO defined as:
\begin{equation}
	ELBO(\theta, \phi) = \sum_{n=1}^{T}\mathbb{E}_{\gamma_{t} \sim q_\phi} [logP_\theta(\boldsymbol{r}_t|\boldsymbol{\gamma}_t)] - KL(q_\phi(\boldsymbol{\gamma}_t|\boldsymbol{\gamma}_{t-1}, \boldsymbol{r}_{1:t})||P_\theta(\boldsymbol{\gamma}_t|\boldsymbol{\gamma}_{t-1}, \boldsymbol{r}_{1:t-1})), 
\end{equation}
and we seek:
\begin{equation}
	\{\theta^*,\phi^*\}=\operatorname*{argmax}_{\theta,\phi}ELBO(\theta,\phi). 
\end{equation}
\subsection{Model Prediction}
Neural GARCH produces 1-step-ahead conditional volatility predictions. Given $\mathcal{I}_{t} = \{\boldsymbol{\gamma}_{t}, \boldsymbol{\Sigma}_{t}, \boldsymbol{r}_{t:t}\}$, we use (\ref{eqn: prediction}) to obtain our prediction of $\boldsymbol{\gamma}_{t+1}$ by drawing from the multivariate normal distribution whose parameters are given by $MLP_{pred}$. We then obtain our estimate of $\boldsymbol{\Sigma}_{t+1}$ deterministically using (\ref{eqn: neural bekk}). To estimate $\boldsymbol{\Sigma}_{t+2}$, we would now have access to $\boldsymbol{r}_{t+1}$ and therefore we obtain the posterior estimate of $\boldsymbol{\gamma}_{t+1}$ using (\ref{eqn: inference}) and predict $\boldsymbol{\Sigma}_{t+2}$ using the posterior estimate of $\boldsymbol{\Sigma}_{t+1}$. This posterior update is crucial as it ensures that we use all available and up-to-date information to predict the next covariance matrix.  
\subsection{Experiments}
We test neural GARCH on a range of daily asset log returns time series covering univariate and multivariate foreign exchange rates (20 pairs), commodity prices (brent crude, silver and gold) and stock indices (DAX, S\&P, NASDAQ, FTSE100, Dow Jones). We provide a brief data description in Table \ref{tab1}. 
\begin{table}[h]
	\centering
	\caption{Description of asset log returns time series analysed in our experiments.}
	\resizebox{0.75\columnwidth}{!}{
		\begin{tabular}{|c|c|c|c|c|}
			\hline
			Dataset & N Time Series & Frequency & Observations & Date Range\\
			\hline
			Foreign exchange & 20 & daily & 3128 & 05/08/2011 - 05/08/2021\\
			\hline
			Brent crude & 1 & daily & 2065 & 05/08/2013 - 05/08/2021\\
			\hline
			Silver \& gold & 2 & daily & 3109 & 05/08/2011 - 05/08/2021\\
			\hline
			Stock indices & 5 & daily & 2054 & 05/08/2013 - 05/08/2021\\
			\hline
	\end{tabular}}
	\label{tab1}
\end{table}

For model training, we split each time series such that 80\% was used in training, 10\% for validation and 10\% for testing. The underyling recurrent neural network (GRU) has a hidden state size 64, the generative and inference MLPs ($MLP_{pred}$ and $MLP_{inf}$) are both 3-layer MLPs with 64 hidden nodes and ReLU activation functions.

For univariate time series, we compare the performance of six models: GARCH(1,1)-Normal, GARCH(1,1)-Student's t, Neural-GARCH(1,1) and Neural-GARCH(1,1)-Student's t, EGARCH(1,1,1)-Normal and EGARCH(1,1,1)-Student's t. Although neural GARCH is an adaptation of the GARCH(1,1) model, we include the EGARCH(1,1,1) model as a benchmark as it is capable of accounting for the asymmetric leverage effect: negative shocks lead to larger volatilities than positive shocks, where the middle index represents the order of the asymmetric term. We would like to investigate whether the data driven approach of neural GARCH allows it to model the leverage effect without the explicit dependence on an asymmetric term as in an EGARCH model. For multivariate time series, we compare the performance of multivariate GARCH(1,1) (BEKK(1,1)) with normal and Student's t innovations against their neural network adaptations. We evaluate the model performance using the log-likelihood of the test dataset. 

\section{Results \& Discussion}
\label{sec: Results}
In Tables \ref{commodity}, \ref{stock index}, \ref{univariate fx} and \ref{multivariate fx} we provide the log-likelihoods evaluated on the test dataset for commodity prices, stock indices, and univariate and multivariate foreign exchange time series. We have highlighed the best model for each time series in bold. For commodity prices, we observe that EGARCH(1,1,1)-Student's t is the best performer on Brent crude, whilst Neural-GARCH(1,1)-Student's t performs best on silver and gold price returns.

For stock indices we observe that Neural-GARCH(1,1)-Student's t performs best on the DAX AND Dow Jones indices whilst EGARCH(1,1,1)-Student's t performs best on S\&P500, NASDAQ and FTSE 100. The fact that the neural GARCH models perform better than EGARCH in some datasets shows that our data-driven approach can learn to accommodate many but not all scenarios of the leverage effect, and therefore in cases where EGARCH outperforms, there are benefits associated with the direct modelling of the asymmetric effect. For univariate foreign exchange time series, we observe that the Neural GARCH variants outperform traditional GARCH models on 16 out of 20 time series, and where neural GARCH outperforms, Neural-GARCH(1,1) with normal innovations performs better on 5/16 time series and Neural-GARCH(1,1)-Student's t performs better on 11/16 time series. 
\begin{table}[h]
	\centering
	\caption{Test log-likelihoods for commodity price time series. Best result highlighed in bold, higher log-likelihood is better.}
	\resizebox{0.85\columnwidth}{!}{
		\begin{tabular}{|c|c|c|c|c|c|c|}
			\hline
			Time series & GARCH(1,1)-Normal & GARCH(1,1)-Student's t & Neural-GARCH(1,1) & Neural-GARCH(1,1)-Student's t & EGARCH(1,1,1)-Normal & EGARCH(1,1,1)-Student's t\\
			\hline
			BRENT &	-298.738 &	-298.689 &	-307.921 &	-295.895 & -299.966 &	$\boldsymbol{-292.798}$\\
			\hline
			SILVER &	-554.595 &	-551.936 &	-541.713 &	$\boldsymbol{-514.476}$ & -572.780 &	-581.834\\
			\hline
			GOLD &	-462.28 &	-450.752 &	-473.074 &	$\boldsymbol{-421.566}$ & -462.857 &	-468.509\\
			\hline
	\end{tabular}}
	\label{commodity}
\end{table}
\begin{table}[h]
	\centering
	\caption{Test log-likelihoods for stock index time series.}
	\resizebox{0.85\columnwidth}{!}{
		\begin{tabular}{|c|c|c|c|c|c|c|}
			\hline
			Time series & GARCH(1,1)-Normal & GARCH(1,1)-Student's t & Neural-GARCH(1,1) & Neural-GARCH(1,1)-Student's t & EGARCH(1,1,1)-Normal & EGARCH(1,1,1)-Student's t\\
			\hline
			DAX &	-261.275 &	-268.944 &	-259.321 &	$\boldsymbol{-244.190}$ & -257.767 &	-266.163\\
			\hline
			SNP &	-300.849 &	-298.614 &	-308.559 &	-295.934 & -300.577 &	$\boldsymbol{-284.841}$\\
			\hline
			NASDAQ &	-327.547 &	-326.401 &	-331.539 &	-320.387 & -334.237 &	$\boldsymbol{-312.366}$\\
			\hline
			FTSE &	-324.437 &	$\boldsymbol{-314.480}$ &	-326.572 &	-315.606 & -322.425 &	$\boldsymbol{-311.135}$\\
			\hline
			DOW &	-298.406 &	-302.196 &	-315.164 &	$\boldsymbol{-284.247}$ & -292.974 &	-293.486\\
			\hline
	\end{tabular}}
	\label{stock index}
\end{table}

For multivariate foreign exchange time series, we observe that Neural-BEKK(1,1)-Student's t is the best performer on 8/9 time series considered. Across different assets we see that the Student's t version of Neural GARCH consistently performs better than the traditional GARCH models as well as Neural GARCH with normal innovations. This suggests that a model with Student's t innovation does indeed model the leptokurtic behaviour of financial time series returns better than a model with normal innovations. This finding is in line with our expectations after surveying the literature (for example \cite{Bollerslev1987} and \cite{Maria2007}). 
\begin{table}[h]
	\centering
	\caption{Test log-likelihoods for univariate foreign exchange time series.}
	\resizebox{\columnwidth}{!}{
		\begin{tabular}{|c|c|c|c|c|c|c|}
			\hline
			Time series & GARCH(1,1)-Normal & GARCH(1,1)-Student's t & Neural-GARCH(1,1) & Neural-GARCH(1,1)-Student's t & EGARCH(1,1,1)-Normal & EGARCH(1,1,1)-Student's t\\
			\hline
			AUDCAD & $\boldsymbol{-397.251}$ & -402.582 & -409.553 & -398.645 & -397.776 &	-473.302\\
			\hline
			AUDCHF & -311.566 & -308.029 & $\boldsymbol{-293.853}$ & -294.010 & -309.295 &	-312.965\\
			\hline
			AUDJPY & -346.024 &	-350.401 &	-353.213 &	$\boldsymbol{-335.945}$ & -346.478 &	-354.095\\
			\hline
			AUDNZD & -303.986 &	-318.345 &	-307.44 &	$\boldsymbol{-301.514}$ & -303.627 &	-322.777\\
			\hline
			AUDUSD & -423.602 &	-424.594 &	-432.518 &	$\boldsymbol{-422.753}$ & -424.498 &	-425.807\\
			\hline
			CADJPY & -351.749 &	-359.545 &	$\boldsymbol{-349.209}$ & -349.842 & -350.460 &	-362.875\\
			\hline
			CHFJPY & -238.566 &	-241.360 &	-215.536 &	$\boldsymbol{-208.710}$ & -230.120 &	-253.050\\
			\hline
			EURAUD & -338.378 &	-344.922 &	-347.995 &	$\boldsymbol{-336.604}$ & -337.481 &	-347.259\\
			\hline
			EURCAD & -347.177 &	-359.499 & 	$\boldsymbol{-345.989}$ & -347.730 & -346.547 &	-366.701\\
			\hline
			EURCHF & -277.643 &	-153.502 &	-156.567 &	$\boldsymbol{-142.963}$ & -275.073 &	-321.051\\
			\hline
			EURGBP & -366.187 &	-378.950 &	-373.515 &	$\boldsymbol{-364.619}$ & -364.727 &	-389.416\\
			\hline
			EURJPY & -266.674 &	-278.327 &	-267.374 &	$\boldsymbol{-256.341}$ & -262.667 &	-290.897\\
			\hline
			EURUSD & -332.917 &	-347.818 &	$\boldsymbol{-330.471}$ & -334.488 & -334.178 &	-361.348\\
			\hline
			GBPAUD & $\boldsymbol{-335.530}$ &	-346.944 &	-353.800 &	-344.842 & -335.812 &	-353.034\\
			\hline
			GBPJPY & -330.030 &	-348.729 &	-337.981 &	$\boldsymbol{-324.559}$ & -329.013 &	-359.506\\
			\hline
			GBPUSD & $\boldsymbol{-418.593}$ &	-431.554 &	-423.460 &	-419.658 & -420.534 &	-441.162\\
			\hline
			NZDUSD & $\boldsymbol{-415.648}$ & 	-416.944 &	-425.841 &	-417.380 & -416.094 &	-417.153\\
			\hline
			USDCAD & -408.008 &	-416.483 &	$\boldsymbol{-404.614}$ &	-413.507 & -406.735 &	-419.863\\
			\hline
			USDCHF & -315.963 &	-303.351 &	-276.461 &	$\boldsymbol{-260.177}$ & -282.682 &	-308.410\\
			\hline
			USDJPY & -295.295 &	-304.539 &	-291.419 &	$\boldsymbol{-277.477}$ & -294.519 &	-318.100\\
			\hline
	\end{tabular}}
	\label{univariate fx}
\end{table}
\begin{table}[h!]
	\centering
	\caption{Test log-likelihoods for multivariate foreign exchange time series.}
	\resizebox{0.85\columnwidth}{!}{
		\begin{tabular}{|c|c|c|c|c|}
			\hline
			Time series & GARCH(1,1)-Normal & GARCH(1,1)-Student's t & Neural-GARCH(1,1) & Neural-GARCH(1,1)-Student's t\\
			\hline
			EURGBP,EURCHF &	-643.521 &	-558.275 &	-523.725 &	$\boldsymbol{-513.214}$\\
			\hline
			GBPJPY GBPUSD &	-629.950 &	-656.198 &	-649.221 &	$\boldsymbol{-605.305}$\\
			\hline
			AUDCHF AUDJPY &	-534.49 &	-522.934 &	-497.726 &	$\boldsymbol{-477.992}$\\
			\hline
			EURGBP,EURUSD,EURJPY &	-920.085 &	-959.420 &	-985.156 &	$\boldsymbol{-917.907}$\\
			\hline
			USDCAD,USDCHF,USDJPY &	-1008.821 &	-998.041 &	-990.601 &	$\boldsymbol{-957.912}$\\
			\hline
			EURGBP,GBPJPY,USDJPY &	$\boldsymbol{-916.957}$ &	-943.66 &	-1011.435 &	-966.806\\
			\hline
			GBPAUD,GBPJPY,GBPUSD &	-971.522 &	-991.8238 &	-1037.296 &	$\boldsymbol{-967.500}$\\
			\hline
			EURCHF,EURGBP,EURJPY,EURUSD &	-1196.477 &	-1127.192 &	-1105.298 & $\boldsymbol{-1078.165}$\\
			\hline
			AUDJPY,AUDCHF,EURCHF,GBPJPY &	-1505.540 &	-862.995 &	-865.471 &	$\boldsymbol{-783.955}$\\
			\hline
	\end{tabular}}
	\label{multivariate fx}
\end{table}

In order to evaluate whether the models' performances across different time series are statistically significant, we plotted a critical difference (cd) diagram by following the approach of the authors in \cite{IsmailFawaz2019} where a Friedman test at $\alpha=0.05$ \cite{Friedman1940} was first used to reject the null hypothesis that the four models are equivalent and have equal rankings, and then a post-hoc test was done using a Wilcoxon signed-rank test \cite{Wilcoxon1945} at the 95\% confidence level. The critical diagram shows average rankings of the models across different datasets. 

\begin{figure}[h!]
	\centering
	\includegraphics[width=0.85\columnwidth,clip,keepaspectratio]{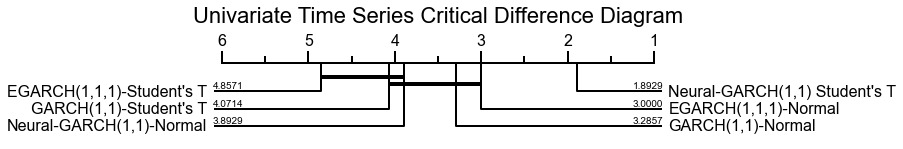}
	\caption{Critical difference diagram of the univariate experiments. A horizontal bold line indicates no significant difference amongst the group of models. We establish that Neural-GARCH(1,1)-Student's t is the best performer in the univariate experiments.}
	\label{cd-univariate}
\end{figure}

In Figure \ref{cd-univariate} we show the cd plot for univariate time series. A bold horizontal line indicates no significant difference amongst the group of models that are on the line. In the univariate experiments we observe no significant difference amongst the group: EGARCH(1,1,1)-Student's T, GARCH(1,1)-Student's T and Neural-GARCH(1,1); likewise, there is also no significant difference amongst the group: GARCH(1,1)-Student's T, Neural-GARCH(1,1), GARCH(1,1)-Normal and EGARCH(1,1,1)-Normal. We also observe that on average, GARCH(1,1)-Normal and EGARCH(1,1,1)-Normal perform significantly better than EGARCH(1,1,1)-Student's T. We establish that Neural-GARCH(1,1)-Student's t is the best performer overall on the univariate datasets, and it significantly outperforms the other models with an average rank of 1.8929.  

\begin{figure}[h!]
	\centering
	\includegraphics[width=0.85\columnwidth,clip,keepaspectratio]{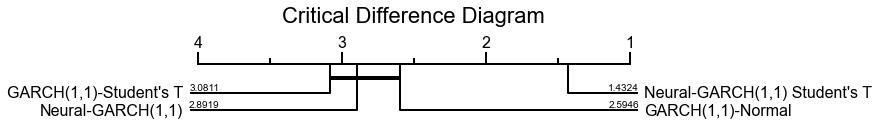}
	\caption{Critical difference diagram showing the average rankings of GARCH(1,1) and Neural-GARCH(1,1) with normal and Student's t innovations on all time series experiments.  We find that Neural-GARCH(1,1)-Student's t is the best-performing model with an average rank of 1.4324.}
	\label{cd}
\end{figure}
In Figure \ref{cd} we show the cd plot constructed using all the time series experiments (univariate and multivariate). Our aim is to compare the class of traditional GARCH(1,1) models against their neural network adaptations. We observe that there is no significant difference between GARCH(1,1)-Student's t, Neural-GARCH(1,1) and GARCH(1,1)-Normal, and we establish that Neural-GARCH(1,1)-Student's t is the best performer overall with an average ranking of 1.4324.

For a GARCH(1,1) model, the returns process is often assumed to be stationary with a constant unconditional mean and variance. Neural GARCH(1,1) relaxes this stationary assumption. The unconditional variance of Neural-GARCH(1,1) in the univariate case
\begin{equation}\label{eq23}
	\sigma^2_t = \omega_t + \alpha_tr^2_t + \beta_t\sigma^2_{t-1}
\end{equation}
is obtained by taking the expectation of (\ref{eq23}):
\begin{equation}
	\begin{split}
		\mathbb{E}[r^2_t] &= \mathbb{E}[\omega_t + \alpha_tr^2_{t-1} + \beta_t\sigma^2_{t-1}]\\
		&= \omega_t + \alpha_t\mathbb{E}[r^2_{t-1}] + \beta_t\mathbb{E}[\sigma^2_{t-1}]\\
		&= \omega_t + (\alpha_t + \beta_t)\mathbb{E}[r^2_{t-1}].
	\end{split}
\end{equation}
For a GARCH(1,1) model with constant coefficients $\{\omega, \alpha, \beta\}$, we have $\mathbb{E}[r^2_t] = \mathbb{E}[r^2_{t-1}]$ (constant unconditional variance)  and therefore $\frac{\omega}{1-\alpha-\beta}$. With Neural-GARCH(1,1), $\mathbb{E}[r^2_t] \neq \mathbb{E}[r^2_{t-1}]$ however we can assume that the parameters $\{\omega_t, \alpha_t, \beta_t\}$ change gradually with no sudden jumps and therefore $\mathbb{E}[r^2_t] \approx \mathbb{E}[r^2_{t-1}]$ \citep{Bringmann2017} and we can approximate the time-varying unconditional variance of Neural-GARCH(1,1) with $\mathbb{E}[r^2_t] \approx \frac{\omega_t}{1-\alpha_t-\beta_t}$ with $\alpha_t + \beta_t < 1$. 

\begin{figure}[h!]
	\centering
	\includegraphics[width=\columnwidth,clip,keepaspectratio]{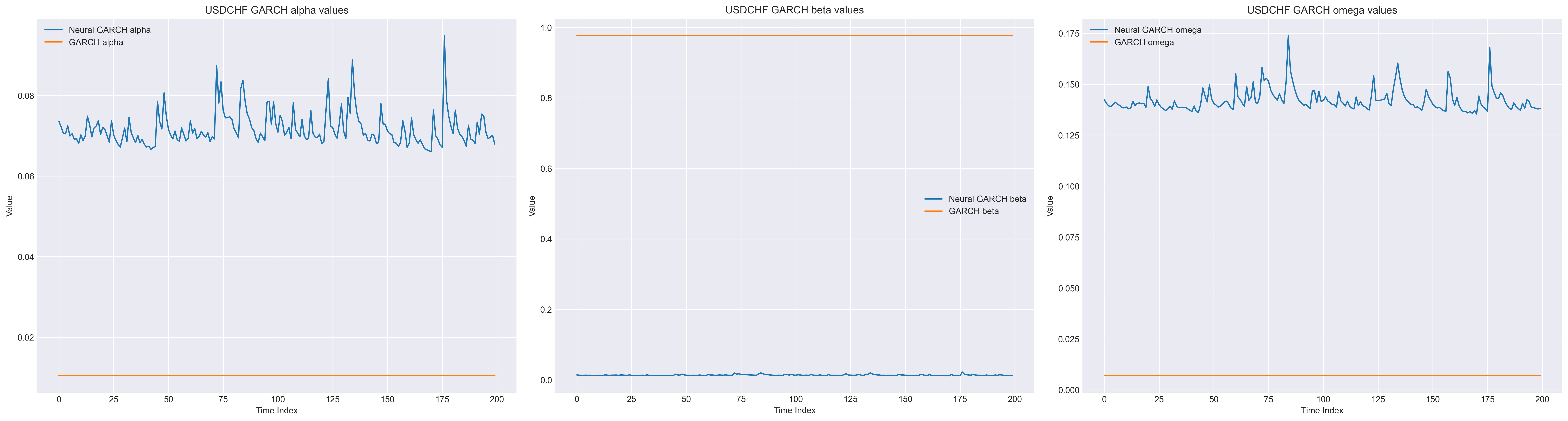}
	\caption{Plots of Neural-GARCH(1,1) coefficients against GARCH(1,1) coefficients. The blue line represents the Neural-GARCH(1,1) $\alpha_t$(left), $\beta_t$(middle) and $\omega_t$(right), and the orange line shows the GARCH(1,1) coefficients.}
	\label{coefficients}
\end{figure}

\begin{figure}[h!]
	\centering
	\includegraphics[width=\columnwidth,clip,keepaspectratio]{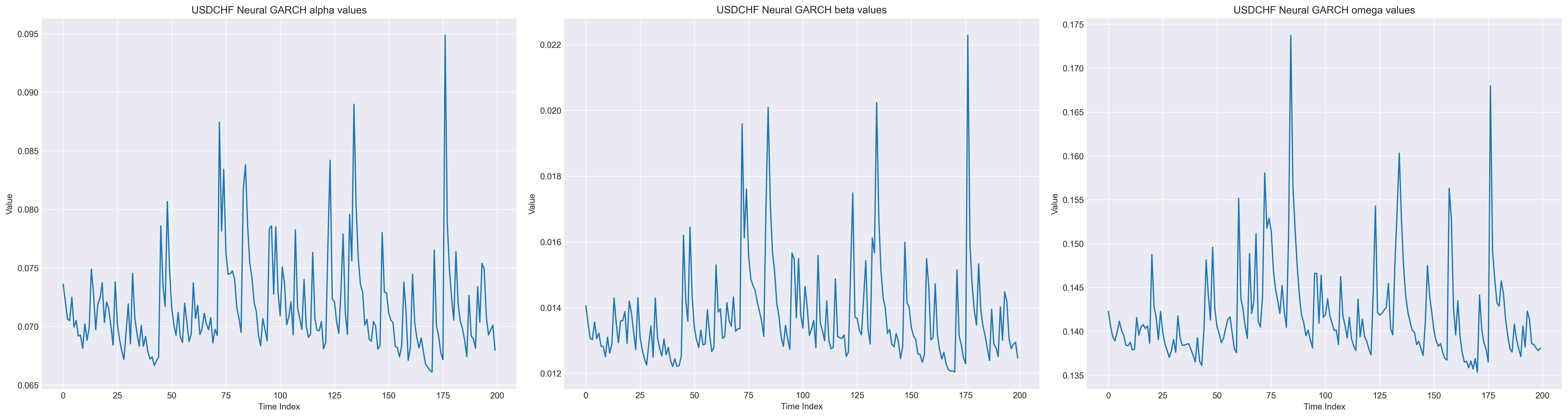}
	\caption{Zoomed-in plots of the Neural-GARCH(1,1) coefficients shown in Figure \ref{coefficients} for USDCHF.}
	\label{coefficientszoom}
\end{figure}

Results from our analysis of the Neural-GARCH(1,1) coefficients show a consistent pattern when compared to GARCH(1,1) models. We provide an example for the currency pair USDCHF in Figure \ref{coefficients}, which shows the time-varying parameter set $\{\omega_t, \alpha_t, \beta_t\}$ of Neural-GARCH(1,1) against the constant set $\{\omega, \alpha, \beta\}$ of GARCH(1,1). We observe across different time series that Neural-GARCH(1,1) consistently estimates a higher value for $\omega$ and $\alpha$, and a lower value for $\beta$. In Figure \ref{coefficientszoom} we show the zoomed-in images of the Neural-GARCH(1,1) coefficients shown in Figure \ref{coefficients} for the currency pair USDCHF. We observe that the coefficients follow well-behaved time-varying behaviour and similar dynamics is observed across all three parameters. This shows the effectiveness of our learned prior neural network ($MLP_{pred}$) which models the distribution $P_\theta(\boldsymbol{\gamma}_t|\boldsymbol{\gamma}_{t-1}, \boldsymbol{r}_{1:t-1})$.

Having time-varying coefficients allows us to model the financial returns time series as a non-stationary process with a 0 unconditional mean but time-varying unconditional variance. Similarly, the authors in \cite{Starica2005} reported that by relaxing the stationarity assumption on daily S\&P 500 returns and using locally stationary linear models, a better forecasting performance was achieved, and in their analysis they showed most of the dynamics of the returns time series to be concentrated in shifts of the unconditional variance. Our model provides a data-driven approach to modelling the returns process. During model training we optimise over the neural network parameters without implementing any external constraints, however we observe in Figure \ref{coefficientszoom} that the model nonetheless outputs time-varying coefficients that satisfy the condition $\alpha_t + \beta_t < 1$, which is required for the model to have a well-defined unconditional variance.

\section{Conclusions}
In this paper we propose neural GARCH: a neural network adaptation of the univariate GARCH(1,1) and multivariate diagonal BEKK(1,1) models to model conditional heteroskedasticity in financial time series. Our model consists of a recurrent neural network that captures the temporal dynamics of the returns process and a multilayer perceptron to predict the next-step-ahead GARCH coefficients, which are then used to determine the conditional volatilities. The generative model of neural GARCH makes predictions based on all available information, and the inference model makes updated posterior estimates of the GARCH coefficients when new information becomes available. We tested two versions of neural GARCH on univariate and multivariate financial returns time series: one with normal innovations and the other with Student's t innovations. When compared against their GARCH counterparts we observe that neural GARCH Student's t is the best performer and from our analysis we hypothesise that this is due to the neural network's ability to capture complex temporal dynamics present in the time series and also allowing us to relax the stationarity assumption that is fundamental to traditional GARCH models. 

\section*{Acknowledgement}{The authors would like to thank Fabio Caccioli, Department of Computer Science, University College London, for proofreading the manuscript and providing feedback.}

\bibliographystyle{rQUF}
\bibliography{sample}
\bigskip
\medskip
\end{document}